\documentclass[runningheads]{llncs}
\usepackage{graphicx}
\usepackage[caption=false]{subfig}
\usepackage{textcomp}
\usepackage{mathtools}
\usepackage{booktabs}
\usepackage{amsmath,amsfonts,amssymb}
\usepackage{chngcntr}
\newtheorem{assumption}{Assumption}

\newcommand\blfootnote[1]{%
  \begingroup
  \renewcommand\thefootnote{}\footnote{#1}%
  \addtocounter{footnote}{-1}%
  \endgroup
}

\begin{document}
\title{MTab: Matching Tabular Data to Knowledge Graph using Probability Models}
\titlerunning{MTab}
%
\author{Phuc Nguyen\inst{1,2} \and
Natthawut Kertkeidkachorn\inst{3} \and \\
Ryutaro Ichise\inst{1,2,3} \and
Hideaki Takeda\inst{1,2}
}
\authorrunning{Phuc Nguyen et al.}
%
\institute{National Institute of Informatics, Japan \and
SOKENDAI (The Graduate University for Advanced Studies), Japan \and National Institute of Advanced Industrial Science and Technology, Japan\\}
\maketitle 
%
\begin{abstract}
\sloppy This paper presents the design of our system, namely MTab, for Semantic Web Challenge on Tabular Data to Knowledge Graph Matching (SemTab 2019). MTab combines the voting algorithm and the probability models to solve critical problems of the matching tasks. Results on SemTab 2019 show that MTab obtains promising performance for the three matching tasks.  
\end{abstract}
%
%
%
\section{Introduction}
\begin{figure}
\centering
\subfloat[\textbf{CTA}]{\label{fig:cta}\includegraphics[width=0.13\textwidth]{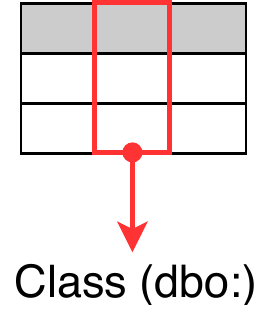}}
\subfloat[\textbf{CEA}]{\label{fig:cea}\includegraphics[width=0.13\textwidth]{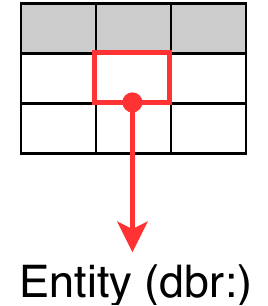}}
\subfloat[\textbf{CPA}]{\label{fig:cpa}\includegraphics[width=0.129\textwidth]{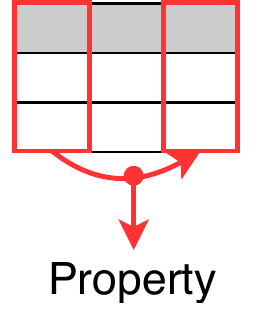}}
\caption{Tabular Data Matching to Knowledge Graph (DBpedia)} 
\label{fig:task}
\end{figure}

Tabular Data to Knowledge Graph Matching (SemTab 2019)\footnote{http://www.cs.ox.ac.uk/isg/challenges/sem-tab/} is a challenge on matching semantic tags from table elements to knowledge graphs (KGs), especially DBpedia. Fig. \ref{fig:task} depicts the three sub-tasks for SemTab 2019. Given a table data, \textbf{CTA} (Fig. \ref{fig:cta}) is the task of assigning a semantic type (e.g., a DBpedia class) to a column. In \textbf{CEA} (Fig. \ref{fig:cea}), a cell is linked to an entity in KG. The relation between two columns is assigned to a property or predicate in KG in \textbf{CPA} (Fig. \ref{fig:cpa}).
\blfootnote{Copyright © 2019 for this paper by its authors. Use permitted under Creative Commons License Attribution 4.0 International (CC BY 4.0).}
\vspace*{-9mm}
\vspace*{-10mm}
\subsection{Problem Definition}
\vspace*{-2mm}
We denotes DBpedia as a knowledge graph $G = (E, T, R)$, where $E, T, R$ are the set of entities, the set of types (or classes), and the set of relations (or predicates) respectively. $e$ is an entity $e \in E$, $t_e$ is the type of an entity  $t_e \in T$, and $r$ is a relation of entity-entity or entity-literal $r \in R$.

Let a table $S$ be a two-dimensional tabular structure consisting of an ordered set of $N$ rows, and $M$ columns. $n_i$ is a row of table ($i = 1...N$), $m_j$ is a column of table ($j = 1...M$). The intersection between a row $n_i$ and a column $m_j$ is $c_{i,j}$ is a value of the cell $S_{i,j}$. In general, the tabular to KG matching problems could be formalized the three sub-tasks as follows.
\begin{itemize}
\item CEA: matching a cell value $c_{i,j}$ into a relevance entity $e \in E$.
\item CTA: matching a column $m_j$ into a exact relevance type and its ancestors.
\item CPA: matching the relation between two columns $m_{j_1}$ and $m_{j_2}$ ($j_1, j_2 \in [1,M], j_1 \ne j_2$) into a relation $r \in R$.
\end{itemize}

\subsection{Assumption} \label{Assumptions}
In MTab, we adopt the following assumptions:
\begin{assumption} \label{closed-world}
MTab is built on a closed-world assumption. It means that the target KG is completed and corrected. 
\end{assumption}
\begin{assumption} \label{Table_types}
The type of input table is vertical relational table. 
\end{assumption}
\begin{assumption} \label{Independece}
Input tables are independence, it means there is no sharing information between input tables. 
\end{assumption}
\begin{assumption} \label{Column}
Column cell values have the same entity types and data types. 
\end{assumption}
\begin{assumption} \label{Header}
The first row of table ($n_1$) is table header. The first cell of column is header of this column, $c_{1,j} \in m_j$.
\end{assumption}
In practice, table headers could have more complicated structures. Headers could available or non-available, be located at the beginning of the table or not, have one rows or multiple rows. In this work, we omit those issues and assume that the table header is located at the first row. 
\section{Approach}
MTab is built on the joint probability distribution of many signals inspired by the graphical probability model-based approach \cite{DBLP:journals/pvldb/LimayeSC10}. However, MTab improves the matching performance by solving two major problems:
\vspace*{-2mm}
\begin{itemize}
    \item Entity lookup: We found that using only DBpedia lookup (the most popular service) does not usually get relevance entities for non-English queries. Therefore, we perform entity lookup on multiple services (with language parameter) to increase the possibility of finding the relevance entities.
    \item Literal matching: We found that mapping cell values to corresponding values in a KG is less effective because the corresponding value in KG is rarely equal with a query value. Therefore, with Assumption \ref{Column}, we adopt literal columns matching to find a relevance relation (property) and aggregate these signals to enhance MTab performance. 
\end{itemize}
Additionally, we also adopt more signals from table elements, introduce a scoring function to estimate the uncertainly from ranking. Note that, the probabilities in this paper could be interpreted as subjective confidences.

\begin{figure}
\centering
\includegraphics[width=1\textwidth]{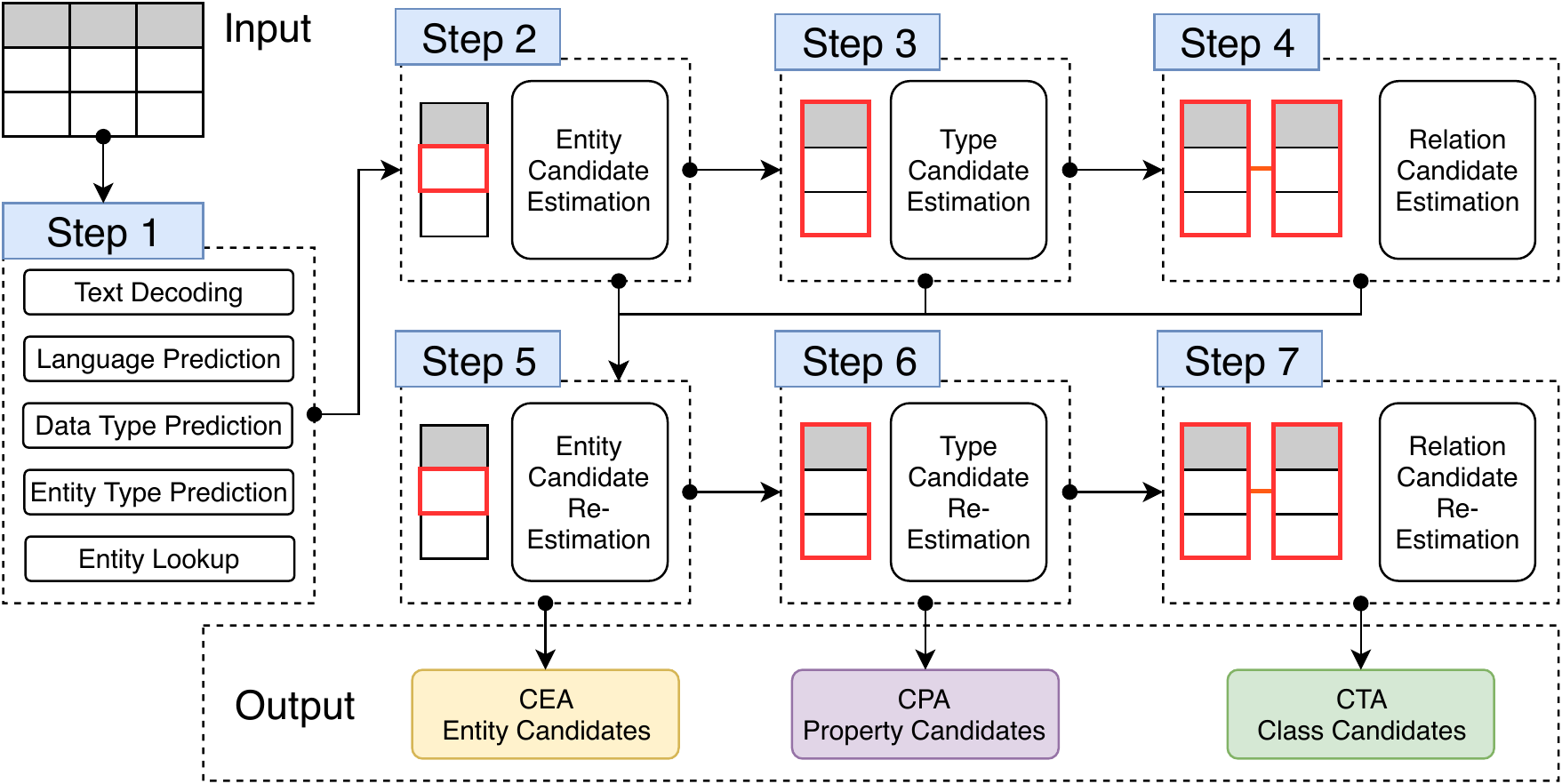}
\caption{The design of MTab framework} 
\label{fig:overall}
\vspace*{-7mm}
\end{figure}
\vspace*{-7mm}
\subsection{Framework}
\vspace*{-2mm}
We design our system (MTab) as 7-steps pipeline (Fig. \ref{fig:overall}). Step 1 is to pre-process a table data $S$ by decoding textual data, predicting languages, data type, entity type prediction, and entity lookup. Step 2 is to estimate entity candidates for each cell. Step 3 is to estimate type candidates for columns. Step 4 is to estimate relationship between two columns. Step 5 is to re-estimate entity candidates with confidence aggregation from step 2, step 3, and step 4. Step 6, and Step 7 are to re-estimate type and relation candidates with results from Step 5, respectively. 
\vspace*{-5mm}
\subsection{Step 1: Pre-processing}
\vspace*{-2mm}
We perform five processes as follows.
\vspace*{-2mm}
\begin{itemize}
    \item \textbf{Text Decoding:} Reading table data has a problem of textual encoding where some characters are loaded as noisy sequences. Loading incorrect encoding might strongly affect the lookup performance, therefore, we used the ftfy tool \cite{speer-2019-ftfy} to fix all noisy textual data in tables.
    \item \textbf{Language Prediction:} We used the pre-trained fasttext models (126 MB) \cite{joulin2017bag} to predict the languages for tables (concatenate all table cell values) and each cell in the table. Since table data is not only written in English but also other languages, determining the language of the input query helpful for the lookup tasks. 
    \item \textbf{Data Type Prediction: } Next, we perform data type prediction to predict 13 pre-defined data types of duckling \footnote{Duckling, link: https://github.com/facebook/duckling} for each cell value in a table $c_{i,j}$. Those types are about numerical tags, email, URL, or phone number. If there is no tag assigned, we assign this cell type as a text tag.
    \item \textbf{Entity Type Prediction: } For each cell value in a table $c_{i,j}$, we also perform entity type prediction with the pre-trained SpaCy models \cite{spacy2} (OntoNotes 5 dataset) to predict 18 entity types. If there is no tag assigned, this cell type is assigned to a text tag. We also manually map from those textual entity types (11 entity types) OntoNotes 5 to some DBpedia classes. 
    \item \textbf{Entity Lookup: } We search the relevance entity on many services including DBpedia Lookup\footnote{DBpedia Lookup, link: https://wiki.dbpedia.org/Lookup}, DBpedia endpoint\footnote{DBpedia Endpoint, link: https://dbpedia.org/sparql}. Also, we search relevant entities on Wikipedia and Wikidata by redirected links to DBpedia to increase the possibility of finding the relevant entities. We use the language information of the table and cell values as the lookup parameters. If there is any non-English lookup URL, it is redirected to the corresponding English URL. We use $\alpha$ \footnote{In MTab, we set $\alpha = 100$} as the limit of lookup results. The search query could be each cell value in a table $c_{i,j}$, or other neighbor cells in the same rows $i$. 
\end{itemize}
\vspace*{-6mm}
\subsection{Step 2: Entity Candidate Estimation}
\vspace*{-2mm}
In this section, we explain how we estimate the entity candidates. Given a cell value $c_{i,j}$, we have a set of ranking result lists from lookup services $Q_{c_{i,j}}$. $q$ is a list of ranking of entities ordered by degree of relevance of a lookup service, where $q \in Q_{c_{i,j}}$. In MTab, we adopted the four services as DBpedia lookup, DBpedia Endpoint, Wikidata lookup, and Wikipedia lookup. However, we can use any services as long as their output is a ranking list of relevance entities. 

Denote $E_{Q_{c_{i,j}}}$ is a set of relevance entities in $Q_{c_{i,j}}$, $s^{q}_{e}$ is a confidence score of an entity $e$ where $e \in E_{Q_{c_{i,j}}}$. The confidence score of entity $e$ is calculated as $s^{Q}_{e} = max(s^{q}_{e})$. $s^{q}_{e}$ is the confidence score of entity $e$ in $q$, $s^{q}_{e} = \alpha - rank_e$, where $rank_e$ is the ranking index of entity in $q$. We normalize those entity confidence score to $[0, 1]$, where $Pr(E_{Q_{c_{i,j}}} | Q_{c_{i,j}}) = 1$, $Pr(e | Q_{c_{i,j}}) = \frac{s^{Q}_{e}}{\sum\limits_{e \in E_{Q_{c_{i,j}}}} s^{Q}_{e}}$ and associate those scores as the potential probability of entities given lookup results.
\vspace*{-5mm}
\subsection{Step 3: Type Candidate Estimation}
\vspace*{-2mm}
Regarding Assumption \ref{Column}, we categorize table columns to entity columns and literal columns. We aggregate the data type (Duckling Tags and SpaCy Tags) from each cell in a column using majority voting. If the majority tag is text or entity-related, the columns is an entity column, else a numerical column. Regarding numerical columns, we perform semantic labeling with EmbNum method \cite{DBLP:conf/jist/NguyenNI018} to annotate relations (DBpedia properties) for numerical columns \footnote{We only use EmbNum for those columns have at least 10 numerical values}. Then, we infer types (DBpedia classes) from those relations. 
\vspace*{-5mm}
\subsubsection{Numerical Column}
The set of numerical columns in table $S$ is $M_{num}$. Given a numerical column $m_j$, we use re-trained EmbNum model on DBpedia \cite{DBLP:conf/jist/NguyenNI018} to derive embedding vector for the list of all numerical values of the column and then search the corresponding relations from the database of labeled attributes \footnote{We used all numerical attributes of DBpedia as the labeled data}. The result $q_{m_j}$ is a ranking of relevance numerical attributes in terms of distribution similarity. We also use $\alpha$ as the limit for ranking result. The confidence score of a relation $r$ is calculated as $s^{m_j}_{r} = \alpha - rank_r$, where $rank_r$ is the ranking index of $r$. These scores are also normalized to a range of [0,1] to associate the probability of potential of relation for numerical columns $Pr(r|m_j)$. 

\begin{figure}
\centering
\includegraphics[width=0.5\textwidth]{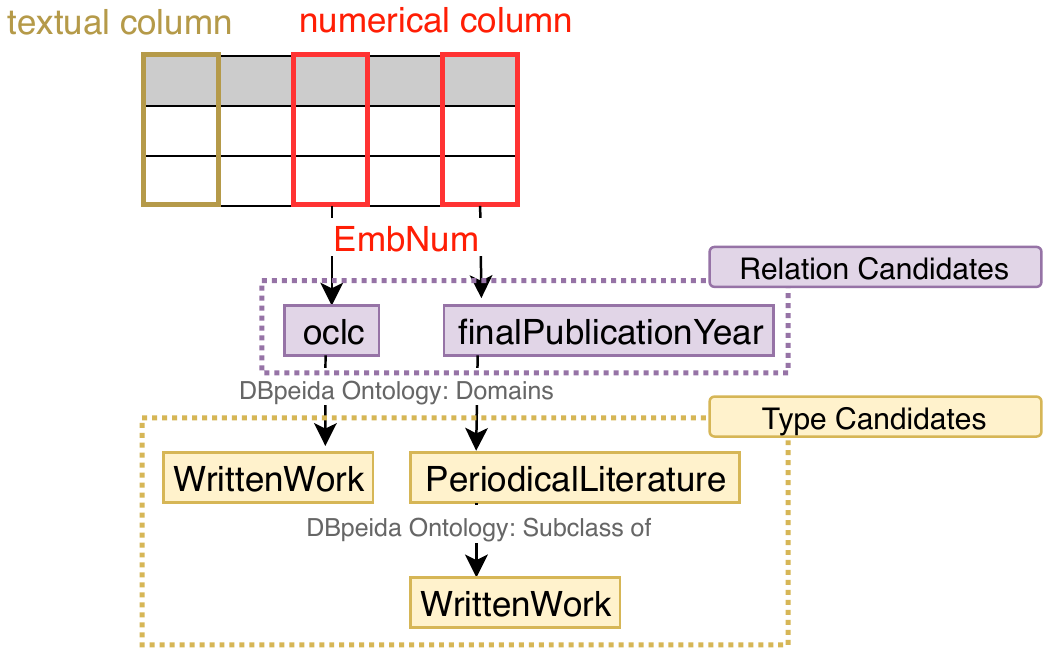}
\caption{Property lookup with EmbNum} 
\label{fig:embnum}
\vspace*{-5mm}
\end{figure}

Next, we use DBpedia endpoint to infer the classes (types) from those relations as Figure \ref{fig:embnum}. $T_{q_{m_j}}$ is a set of inferred types, $t$ denotes a type $t \in T_{q_{m_j}}$. Those types will be used for entity columns. The confidence score of types is estimated as $s_{{r}_{t}} = max(s^{M_{num}}_{r})$. Then, we normalized those scores to [0,1] so that $Pr(T_{q_{M_{num}}}) = 1$, those confidence scores are associated as the probabilities of type potential $Pr(t|M_{num})$ given $M_{num}$.
\vspace*{-3mm}
\vspace*{-5mm}
\subsubsection{Entity Column}
Given a set of entity columns in table $S$ is $M_{ent}$, we consider these signals from
\begin{enumerate}
\sloppy \item $Pr(t|M_{num})$: the probabilities of type potential from numerical columns
\sloppy \item $Pr(t|m_j, Q_{m_j})$: the probabilities of type potential aggregated from the types of entity lookup for the all cells in column $m_j$ ( $Pr(t|m_j, Q_{m_j}) = \sum\limits_{c_{i,j} \in m_j} Pr(t|Q_{c_{i,j}}) $ ). We normalized these aggregated potentials and associates these as potential probabilities. 

\item $Pr(t|m_j, SpaCy_{m_j})$: the probabilities of type potential aggregated from SpaCy entity type prediction for the all cell in column $m_j$. We used majority voting and normalized these voting value to [0,1]. Then, we associate those normalized voting value type potential probabilities. 

\item $Pr(t|c_{1,j})$: the probabilities of type potential given header value of the column $m_j$. We associate the normalized Levenshtein distance as potential probability that a type (DBpedia class) correspond with a header value. 
\end{enumerate}
The probabilities of type potential is derived from the four signals as $Pr(t|m_j) = w_1 Pr(t|M_{num}) w_2 Pr(t|m_j, Q_{m_j}) w_3 Pr(t|m_j, SpaCy_{m_j}) w_4 Pr(t|c_{1,j})$, where $w_1,\\ w_2,\ w_3,\ w_4$ are learnable weights. Note that, some probabilities of signals might be 0 or too small, and aggregate those might add too much noise to the final aggregation. Therefore, if any signal probabilities less than $\beta$\footnote{In MTab, $\beta = 0.5$},  we omit those signals.
After aggregation, we also perform normalization for $Pr(t|m_j)$ to a range of [0,1] so that $Pr(T_{m_j}|m_j) = 1$.
\vspace*{-5mm}
\subsection{Step 4: Relation Candidate Estimation}
\vspace*{-2mm}
Given two columns $m_{j_1}$ and $m_{j_2}$, we estimate the probabilities of relation potential of $Pr(r|m_{j_1},m_{j_2})$. We consider two type of relation between two columns: Entity column to Entity column and Entity column to non-Entity column. To be simple, we associate the first entity column is $m_{j_1}$. If the second column is entity column, we denote it as $m^{ent}_{j_2}$, else $m^{non-ent}_{j_2}$. 
\vspace*{-5mm}
\subsubsection{Entity - Entity columns $Pr(r|m_{j_1}, m^{ent}_{j_2})$:} 
Given $c_{i,j_1}$ is a cell value of the column $m_{j_1}$ and the row $r_{i}$, $c_{i,j_2}$ is a cell value of the column $m^{ent}_{j_2}$. We assume that there is a relation between entity candidates of $c_{i,j_1}$ and $c_{i,j_2}$, therefore we use DBpedia endpoint to find how many links (relations or properties) between entity candidates of $c_{i,j_1}$ and $c_{i,j_2}$. The confidence score of relation is calculated as
$s^{c_{i,j_1}, c_{i,j_2}}_r = 1$ if there is any relation between entity candidates of two columns. Then, we aggregate those scores of all rows to get the candidate score for two columns as $s^{m_{j_1}, m^{ent}_{j_2}}_r = \sum\limits_{i \in [1,N]} s^{c_{i,j_1}, c_{i,j_2}}_r $. Then, we normalize those score to a range of [0,1] so that $Pr(R_{m_{j_1}, m^{ent}_{j_2}}|m_{j_1}, m^{ent}_{j_2})$ and associate it as the probability of relation potential of Entity and Entity Columns $Pr(r|m_{j_1}, m^{ent}_{j_2})$.
\vspace*{-5mm}
\subsubsection{Entity - Non-Entity columns $Pr(r|m_{j_1}, m^{non-ent}_{j_2})$:} 
\sloppy Given $c_{i,j_1}$ is a cell value of the column $m_{j_1}$ and the row $r_{i}$, $c_{i,j_2}$ is a cell value of the column $m^{non-ent}_{j_2}$. We estimate the relevance ratio between entity candidates and non-entity value $c_{i,j_2}$. Given an entity candidate $e$ have pairs of relation($r_e$)-values($v_e$), we compare the non-entity value $c_{i,j_2}$ with all attribute values $v_e$. We select those pairs have ratio larger than $\beta$. We only compare two values of $c_{i,j_1}$ and $v_e$ based on there data types (textual type or numerical type).
\begin{itemize}
    \item For textual values: We use the normalized Levenshtein distance to estimate the relevance ratio between $v_e$ and $c_{i,j_2}$ as $s(v_e, c_{i,j_2})$.
    \item For numerical values: the relevance ratio is calculated as
\end{itemize}
\begin{equation}
        s(v_e, c_{i,j_2}) = 
        \begin{cases}
        0, & \text{if}\ max(|c_{i,j_2}|,|v_e|) = 0 \text{ and}\ |c_{i,j_2}- v_e|\ne 0 \\ 
        1, & \text{if}\ max(|c_{i,j_2}|,|v_e|) = 0 \text{ and}\ |c_{i,j_2}- v_e| = 0 \\
        1 - \frac{|c_{i,j_2} - v_e|}{max(|c_{i,j_2}|,|v_e|)} , & \text{if}\ max(|c_{i,j_2}|,|v_e|)  \ne 0 \\
        \end{cases}
\end{equation}
We aggregate all relevance ratio with respect to relations. Then we normalize those aggregated ratio to [0,1], and associate this as probability of relation potential. 
$Pr(r|m_{j_1}, m_{j_2})$. If the column of $m_{j_2}$ is numerical columns, we also aggregate the re-calculated probability from $Pr(r|m_{j_2})$ (step 3) as:
\begin{equation}
    Pr(r|m_{j_1}, m_{j_2}, m_{j_2} \text{is numerical}) = w_5 Pr(r|m_{j_1}, m_{j_2}) w_6 Pr(r|m_{j_2})
\end{equation}
where $w_5, w_6$ are learnable parameters.
\vspace*{-5mm}
\subsection{Step 5: Entity candidate Re-Estimation}
\vspace*{-2mm}
In this step, we present a method to re-estimate the probabilities of entity candidates $Pr(e|S)$. Given a cell $S_{i,j}$ containing a cell value $c_{i,j}$ at row $n_i$, and column $m_j$, we consider these signals from:
\begin{itemize}
    \item $Pr(e|Q_{c_{i,j}})$: The entity candidate probabilities given look up results.
    \item $Pr(e|m_j)$: The probabilities of entity candidates given their type's probabilities (Step 3). This can be estimated by $Pr(e|m_j) = max(Pr(t_e|m_j, Q_{m_j}))$, where $t_e$ is a type of the entity $e$.
    \item $Pr(e|c_{i,j})$: The probabilities of entity candidates given the cell value $c_{i,j}$. We get the mean ratio of the normalized Levenshtein distance, heuristic abbreviation rules (first character of words, titles, dates, time).
    \item $Pr(e|n_i,m_{j_1})$: The probabilities of entity candidates given cell values in a row $c_{i,j} \in n_i$. We do the same procedure as Step 4 to compare all entity values with a cell value, and compute the mean probability for all cell value in a row. $Pr(e|n_i,m_{j_1}) = mean(Pr(e|m_{j_1}, m_{j_2}))$, where $j_1 \ne j_2$.
\end{itemize}
Overall, the equation is as follows.
\begin{equation}
    Pr(e|S) = w_7 Pr(e|Q_{c_{i,j}}) w_8 Pr(e|m_j) w_9 Pr(e|c_{i,j}) w_{10} Pr(e|n_i,m_{j_1})
\end{equation}
where $w_7, w_8, w_9, w_{10}$ are learnable parameters.
\vspace*{-4mm}
\subsection{Step 6, 7: Re-Estimate Types and Relations}
\vspace*{-2mm}
We select the highest probabilities of entity candidates in Step 5 for each cell $S_{i,j}$ to re-estimate types and relations with majority voting. 
\vspace*{-5mm}
\section{Results}
\vspace*{-2mm}
Table \ref{tab:res} reports the overall results (The Primary Score) of MTab for three matching tasks in the four rounds of SemTab 2019 \cite{hassanzadeh_oktie_2019_3518539}. Overall, these results show that MTab achieves promising performances for the three matching tasks. 

\begin{table}[!ht]
\vspace*{-5mm}
\caption{Overall result of MTab system on the four round dataset of SemTab 2019}
\label{tab:res}
\centering
\setlength{\tabcolsep}{2.em} 
{\renewcommand{\arraystretch}{1.1}
\begin{tabular}{@{}|l|c|c|c|@{}}
\hline
\textbf{SemTab}  & \textbf{CEA (F1)} & \textbf{CTA (AH)}   & \textbf{CPA (F1)} \\ \hline
\textbf{Round 1} & 1.000        & 1.000 (F1) & 0.987    \\
\textbf{Round 2} & 0.911    & 1.414      & 0.881    \\
\textbf{Round 3} & 0.97     & 1.956      & 0.844    \\
\textbf{Round 4} & 0.983    & 2.012      & 0.832    \\ \hline
\end{tabular}
}
\end{table}
\footnotetext{https://github.com/phucty/MTab} 
\vspace*{-7mm}
\section{Discussion}
\vspace*{-2mm}
In this paper, we present MTab for the three matching tasks in SemTab 2019.

\textbf{Novelty: }The MTab performance might be explained in part by tackle the two major problems of the three matching tasks. MTab is built on top of multiple lookup services, therefore, it increases the possibility of finding the relevant entities. Additionally, MTab adopted many new signals (literal) from table elements and use them to enhance matching performance. 

\textbf{Limitation: } Since MTab is built on the top of lookup services, therefore, the upper bound of accuracy strongly relies on the lookup results. In MTab, it is computation-intensive because of aggregating the confidence signals from many parts of the table. Therefore, MTab is not suitable for the real-time application, where we need to get the result as fast as possible. MTab could be modified to match only some parts of the table to reduce the processing time as Table Miner+ \cite{zhang2017effective}. However, we find that this is a trade-off between effectiveness and efficiency when using Table Miner+ \cite{zhang2017effective} method. A concrete analysis of the trade-off issue is left as our future investigation.

\textbf{Future work: }
MTab could be improved by relaxing our assumptions:
\vspace{-2mm}
\begin{itemize}
    \item Assumption \ref{closed-world}: The closed-world assumption might  not hold in practice. Improving the completeness and correctness of knowledge graph might improve MTab performance.
    \item Assumption \ref{Table_types}: Classify table types before matching could help to improve MTab performance \cite{Nishida2017}. 
    \item Assumption \ref{Independece}: In reality, some tables could have shared schema. For example, tables on the Web could be divided to many web pages, therefore we can expect improving matching performance by stitching those tables in the same web page (or domain) \cite{Lehmberg:2017:SWT:3137628.3137657}, \cite{RitzePhd}. Therefore, performing holistic matching could help improve  MTab performance.
    \item Assumption \ref{Header}: Correctly recognize table headers could help to improve MTab performance.
\end{itemize}
\vspace*{-5mm}
\bibliographystyle{unsrt}
\bibliography{cite}
\end{document}